\documentclass[runningheads]{llncs}
\usepackage{graphicx}
\usepackage{comment}
\usepackage{amsmath,amssymb} % define this before the line numbering.
\usepackage{color}

\usepackage{epsfig}
\usepackage{caption}
\usepackage{verbatim}
\usepackage{pifont}
\usepackage{booktabs}  
\usepackage{subfigure}
\usepackage{float}
\usepackage{url}
\usepackage{wrapfig}
\usepackage{xspace}

\usepackage{pifont}% http://ctan.org/pkg/pifont
\newcommand{\xmark}{\ding{55}}%

\usepackage[hidelinks]{hyperref}

\hypersetup{
    colorlinks=true,
    linkcolor=red,
    filecolor=magenta,
    urlcolor=magenta,
    citecolor=green,
}

\begin{document}
% \linenumbers
\pagestyle{headings}
\mainmatter
\def\ECCVSubNumber{2252}  % Insert your submission number here

% abbreviation
\makeatletter
\DeclareRobustCommand\onedot{\futurelet\@let@token\@onedot}
\def\@onedot{\ifx\@let@token.\else.\null\fi\xspace}
\def\eg{\emph{e.g}\onedot} 
\def\Eg{\emph{E.g}\onedot}
\def\ie{\emph{i.e}\onedot} 
\def\Ie{\emph{I.e}\onedot}
\def\cf{\emph{c.f}\onedot} 
\def\Cf{\emph{C.f}\onedot}
\def\etc{\emph{etc}\onedot} 
\def\vs{\emph{vs}\onedot}
\def\wrt{w.r.t\onedot} 
\def\dof{d.o.f\onedot}
\def\etal{\emph{et al}\onedot}

\title{Boundary-preserving Mask R-CNN} % Replace with your title

\begin{comment}
\titlerunning{ECCV-20 submission ID \ECCVSubNumber} 
\authorrunning{ECCV-20 submission ID \ECCVSubNumber} 
\author{Anonymous ECCV submission}
\institute{Paper ID \ECCVSubNumber}
\end{comment}
%******************

%\begin{comment}
\titlerunning{Boundary-preserving Mask R-CNN}
\author{Tianheng Cheng\inst{1} \and
        Xinggang Wang$^\dag$\inst{1}  \and
        Lichao Huang\inst{2}   \and
        Wenyu Liu\inst{1}}
\authorrunning{Cheng et al.}
\institute{Huazhong University of Science and Technology \\
\email{\{thch,xgwang,liuwy\}@hust.edu.cn}\\
\and
Horizon Robotics Inc. \\
\email{lichao.huang@horizon.ai}}
%\end{comment}
%******************
\maketitle

\let\thefootnote\relax\footnotetext{$^\dag$Xinggang Wang is the corresponding author.}

\begin{abstract}
  Tremendous efforts have been made to improve mask localization accuracy in instance segmentation. 
  Modern instance segmentation methods relying on fully convolutional networks perform pixel-wise classification, 
  which ignores object boundaries and shapes, leading coarse and indistinct mask prediction results and imprecise localization. 
  To remedy these problems, we propose a conceptually simple yet effective Boundary-preserving Mask R-CNN (BMask R-CNN) to leverage object boundary information to improve mask localization accuracy. 
  BMask R-CNN contains a boundary-preserving mask head in which object boundary and mask are mutually learned via feature fusion blocks. As a result, the predicted masks are better aligned with object boundaries. 
  Without bells and whistles, BMask R-CNN outperforms Mask R-CNN by a considerable margin on the COCO dataset; in the Cityscapes dataset, there are more accurate boundary groundtruths available, so that BMask R-CNN obtains remarkable improvements over Mask R-CNN. 
  Besides, it is not surprising to observe that BMask R-CNN obtains more obvious improvement when the evaluation criterion requires better localization (\eg, AP$_{75}$) as shown in Fig.~\ref{fig:ap_versus_iou}. Code and models are available at \url{https://github.com/hustvl/BMaskR-CNN}.

\end{abstract}

\section{Introduction}
Instance segmentation, a fundamental but challenging task in computer vision, aims to assign a pixel-level mask to localize and categorize each object in images, driving numerous vision applications such as autonomous driving, robotics and image editing. With the rapid development of deep convolutional neural networks (DCNN), various methods based on DCNN were proposed for instance segmentation. Prevalent methods for instance segmentation are based on object detection, which provides box-level localization information for instance-level segmentation, among which Mask R-CNN~\cite{HeGDG17} is the most successful one. It extends Faster R-CNN~\cite{RenHG017} by adding a simple fully convolutional network (FCN) to predict the mask of each detected instance. Due to the great effectiveness and flexibility, Mask R-CNN serves as a state-of-the-art baseline and has facilitated most recent instance segmentation research, such as \cite{HuangHGHW19,LiuQQSJ18,ChenPWXLSF0SOLL19,CaiNArxiv19,abs-1810-10327}.

\begin{figure}[t]
    \centering
    \begin{minipage}{.45\textwidth}
    \centering
    \includegraphics[width=1.1\linewidth]{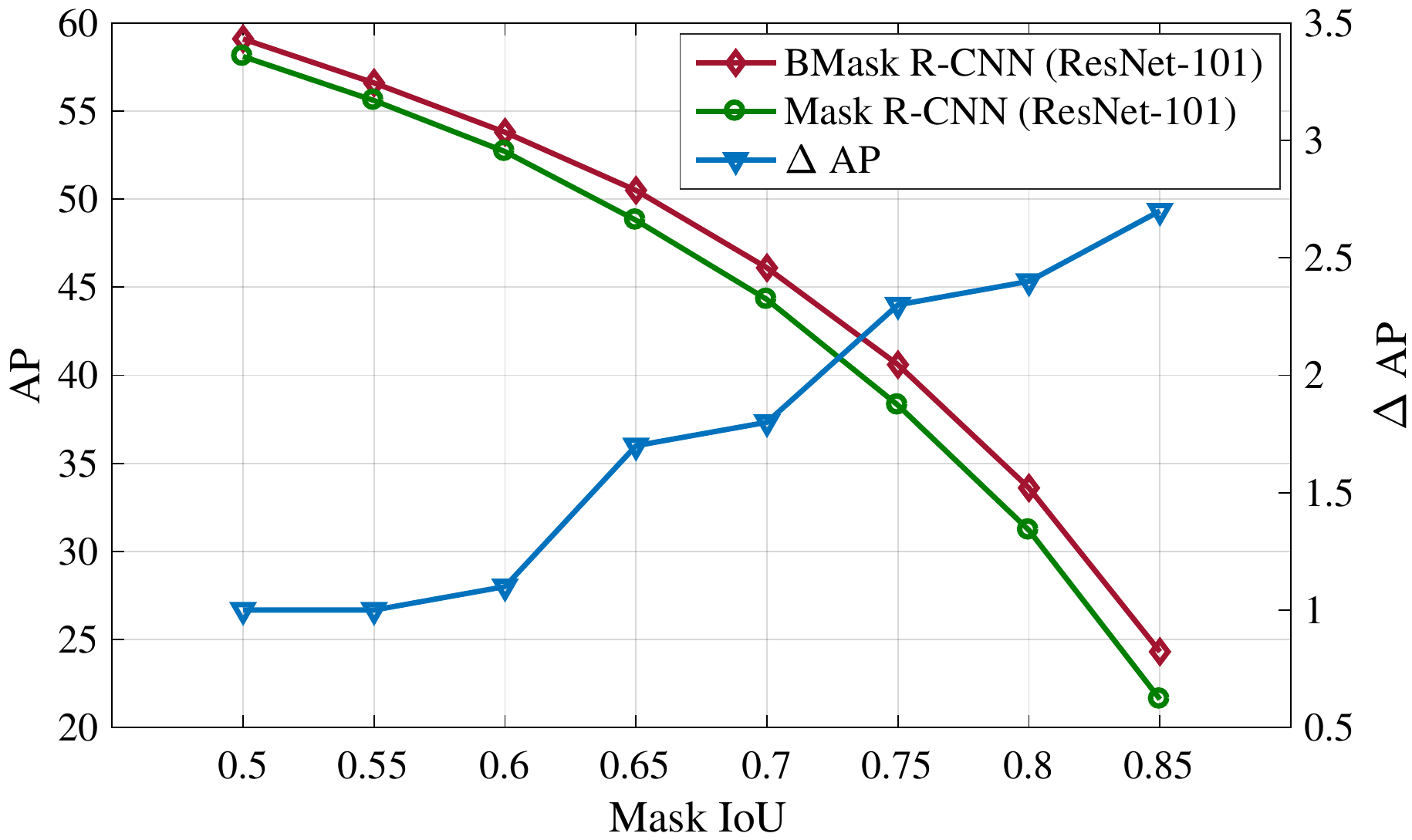}
    \caption{AP curves of Mask R-CNN and BMask R-CNN under different mask IoU thresholds on the COCO \textit{val2017} set. The blue line shows the AP gains of BMask R-CNN over Mask R-CNN.}
    \label{fig:ap_versus_iou}
    \end{minipage}%
    \hspace{0.05\textwidth}
    \begin{minipage}{0.45\textwidth}
    \centering
    \includegraphics[width=\linewidth]{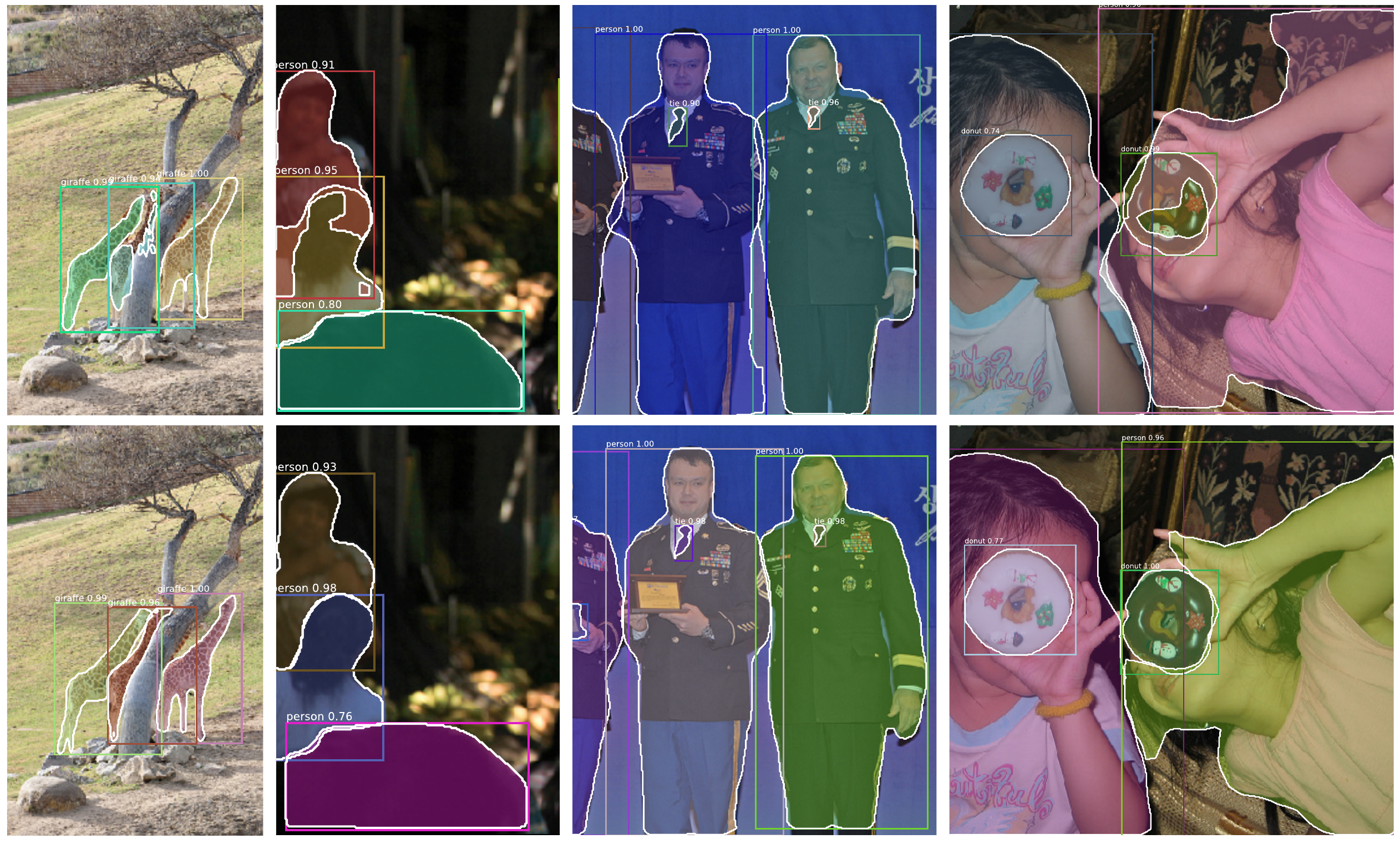}
    \caption{\textbf{First row}: Selected cases of coarse boundaries appeared in the instance segmentation results of Mask R-CNN. \textbf{Second row}: Our proposed method can predict more precise boundaries.}
    \label{fig:comparison_w_wo_edge}
    \end{minipage}
    \vspace{-0.4cm}
\end{figure}

\begin{comment}
\begin{figure}[tp]
    \centering
    \includegraphics[width=\linewidth]{figures/ap_versus_iou.pdf}
    \caption{AP curves of Mask R-CNN and BMask R-CNN under different mask IoU thresholds on the COCO \textit{val2017} set. The blue line shows the AP gains of BMask R-CNN over Mask R-CNN.}
    \label{fig:ap_versus_iou}
\end{figure}  
\end{comment}

In the Mask R-CNN framework, state-of-the-art instance segmentation networks \cite{HeGDG17,HuangHGHW19,LiuQQSJ18} obtain instance masks by performing pixel-level classification via FCN. It treats all pixels in the proposal equally and ignores the object shape and boundary information. However, the pixels near boundaries are hard to be classified. Evidently, it is hard for pixel-level classifier to guarantee precise masks. We find that fine boundaries can provide better localization performance and make the object masks more distinct and clear. As illustrated in Fig.~\ref{fig:comparison_w_wo_edge}, Mask R-CNN (the first row) without consideration about boundaries is prone to output coarse and indistinct segmentation results with unreasonable overlaps between objects in comparison with the one that involves boundaries (the second row). 

To address this issue, we leverage instance boundary information to enhance the mask prediction. Instance boundary is a dual representation of instance mask and it can guide the mask prediction network to output masks that are well-aligned with their groundtruths. Thus, the masks are more distinct and give more precise object location. Based on this motivation, we propose a conceptually simple and novel Boundary-preserving Mask R-CNN (BMask R-CNN) that unifies instance-level mask prediction and boundary prediction in one network. 

Specifically, based on Mask R-CNN, we replace the original mask head with the proposed boundary-preserving mask head which contains two sub-networks for jointly learning object masks and boundaries. We insert two feature fusion blocks to strengthen the connection between boundary feature learning and mask feature learning. At last, mask prediction is guided by boundary features which contain abundant shape and localization information. The main purpose of learning boundaries is to capture features for precise object localization. Nevertheless, learning boundaries is non-trivial, because boundary groundtruths that generated from sparse annotated polygons ($\ie$, in the COCO dataset \cite{LinMBHPRDZ14}) are noisy and boundary classification has less training pixels than that for mask classification. To solve this problem, we further dive into the optimization for boundary learning by performing studies about boundary loss and exploit a boundary classification loss by combining binary cross-entropy loss and the dice loss~\cite{MilletariNA16}.

We perform extensive experiments to evaluate the performance of BMask R-CNN. On the challenging COCO dataset~\cite{LinMBHPRDZ14}, BMask R-CNN achieves considerably significant improvements compared with Mask R-CNN regardless of the backbones. Note that our BMask R-CNN provides larger gains if it requires more precise mask localization, as shown in Fig.~\ref{fig:ap_versus_iou}. On the fine-annotated Cityscapes dataset~\cite{CordtsORREBFRS16}, BMask R-CNN brings larger improvements with better mask annotations. 

The main contributions of this paper can be summarized as follows.
\begin{itemize}
    \item We present a novel Boundary-preserving Mask R-CNN (BMask R-CNN), which is the first work that explicitly exploits object boundary information to improve mask-level localization accuracy in the state-of-the-art Mask R-CNN framework.
    \item BMask R-CNN is conceptually simple yet effective. Without bells and whittles, BMask R-CNN outperforms Mask R-CNN by $1.7\%$ AP and $2.2\%$ AP on the COCO val set and the Cityscapes test set respectively. Further, BMask R-CNN obtains higher AP gains when the mask IoU threshold becomes higher, as shown in Fig.~\ref{fig:ap_versus_iou}. 
    \item We perform ablation studies on the components of BMask R-CNN, \eg, feature fusion blocks, boundary features, boundary losses and the Sobel mask head, which are helpful to interpret how BMask R-CNN works and provide some thoughts for further research on instance segmentation.
\end{itemize}

\section{Related Work}

\subsubsection{Instance Segmentation:} 
Existing methods can be divided into two categories, \ie detection-based methods and segmentation-based methods. Detection-based methods employ object detectors~\cite{Girshick15,RenHG017,DaiLHS16,LinDGHHB17} to generate region proposals and then predict their masks after RoI pooling/align~\cite{Girshick15,HeGDG17}. Based on CNN, % DeepMask~\cite{PinheiroCD15}, SharpMask and InstanceFCN~\cite{DaiHLR016} 
\cite{PinheiroCD15,DaiHLR016,PinheiroLCD16} predict masks for object proposals. 
FCIS~\cite{LiQDJW17} extends InstanceFCN by exploiting position-sensitive inside/outside score maps and fully convolutional networks for instance segmentation.
MNC~\cite{DaiHS16} using end-to-end cascade networks divides instance segmentation into three tasks: regress bounding boxes, estimate instance masks and categorize instances.  
BAIS~\cite{HayderHS17} uses boundary-based distance transform to predict mask pixels that are beyond bounding boxes. 
Mask R-CNN~\cite{HeGDG17} extends Faster R-CNN~\cite{RenHG017} by adding a mask prediction branch in parallel with the existing box regression and classification branches, demonstrating competitive performance on both object detection and instance segmentation. 
PANet~\cite{LiuQQSJ18} based on Mask R-CNN introduces the bottom-up path augmentation for FPN~\cite{LinDGHHB17} to enhance information flow and adaptive feature pooling for better mask features. 
MaskLab~\cite{ChenHPS0A18}, built on Faster R-CNN, obtains instance-level masks by combining cropped semantic segmentation and direction prediction which separate objects of different semantic classes and same semantic classes respectively. Mask scoring R-CNN~\cite{HuangHGHW19} addresses the misalignment  between mask quality and mask score in Mask R-CNN by explicitly learning the quality of predicted masks. \cite{ChenPWXLSF0SOLL19} further improves Cascade Mask R-CNN~\cite{CaiNArxiv19} by interweaving box and mask branches in a multi-stage cascade manner and providing spatial context through semantic segmentation. 
Huang~\etal apply a criss-cross attention module~\cite{huang2020ccnet} to capture the full-image contextual information for instance segmentation.
\cite{kirillov2019pointrend} draws on the idea of rendering and adaptively selects key points to recover fine details for high-quality image segmentation. 

Segmentation-based methods first exploit pixel-level segmentation over the image and then group the pixels together for each object. InstanceCut~\cite{KirillovLASR17} adopts boundaries to partition semantic segmentation into instance-level segmentation. 
SGN~\cite{LiuJFU17} groups pixels along rows and columns by line segments. \cite{UhrigCFB16} utilizes predicted instance centers and pixel-wise directions to group instances with fully convolutional networks. Recently, several methods~\cite{abs-1708-02551,FathiWRWSGM17} take the advantage of deep metric learning to learn the embedding for instances and then group pixels to form instance-level segmentation.

% \vspace{-0.3cm}
\subsubsection{Boundary, Edge and Segmentation:}
Deep fully convolutional neural networks has achieved great progress in edge detection. 
Xie~\etal propose the fully convolutional holistically-nested edge detector HED~\cite{XieT17} which performs in an image-to-image manner and end-to-end training. 
CASENet~\cite{YuF0R17} presents a novel challenging task semantic boundary detection, aiming to detect category-aware boundaries. 
\cite{YuLZFRKK18,AcunaKF19} investigate the label misalignment problem caused by noisy labels in semantic boundary detection. 
\cite{xu2019} proposes geometric aware loss function for object skeleton detection in nature images. In semantic segmentation, Chen~\etal~\cite{ChenPKMY18} propose fully connected conditional random field (CRF)~\cite{KrahenbuhlK11} to capture spatial details and refine boundaries. 
Recent semantic segmentation methods~\cite{ChenBP0Y16,YuWPGYS18,takikawa2019gated,BertasiusST16,HuangXZSXK16} leverage predicted boundaries or edges to facilitate semantic segmentation. \cite{ChengCWSCTZX2020,YuanXCW20} refine segmentation results with direction fields learned from predicted boundaries.
% While using predicted boundaries to refine mask predictions brings minor improvements in instance segmentation. 
Zimmermann~\etal propose edge agreement head~\cite{abs-1809-07069} to focus on boundaries of instances with an auxiliary edge loss. Different from these previous methods, BMask R-CNN explicitly predicts instance-level boundaries, from which we obtain instance shape information for better mask localization.
Compared to semantic segmentation, boundaries in instance segmentation have dual relations to the masks. Therefore, we build fusion blocks to mutually learn boundary and mask features and improve the representations for mask localization and lead the mask prediction focus more on the boundaries. 

\section{Boundary-preserving Mask R-CNN}

\subsection{Motivation}

\begin{figure}[htb]
    \centering
    \includegraphics[width=0.5\linewidth]{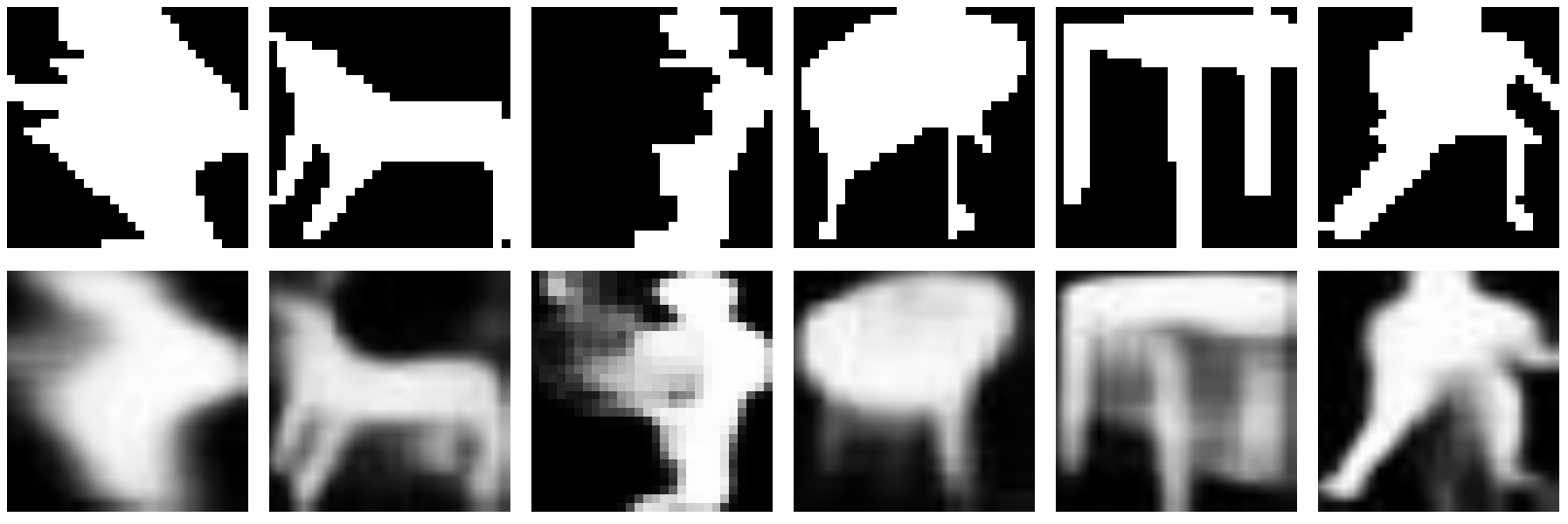}
    \caption{Visualization of some predicted masks (in the bottom row) of Mask R-CNN \vs their groundtruths (in the top row).}
    \label{fig:motivation_figure}
\end{figure}

In Mask R-CNN, instance segmentation is performed based on pixel-level predictions. To learn a translation invariant predictor, predictions are made based on the local information. Though the local features extracted using deep network have large receptive fields, the shape information of object is ignored. Thus, the predicted masks often contain coarse and indistinct as well as some false positive predictions.
For better understanding this problem, we analyze and visualize some raw mask prediction from Mask R-CNN with ResNet-50~\cite{HeZRS16} and FPN. As shown in Fig.~\ref{fig:motivation_figure}, some mask predictions are rough and imprecise.
Obviously, employing object boundaries will be helpful to address this issue by providing better localization and guidance. Therefore, we propose a Boundary-preserving Mask R-CNN to exploit boundary information to guide more precise mask prediction.

\vspace{-1cm}
\begin{figure}[htb]
    \centering
    \includegraphics[width=\linewidth]{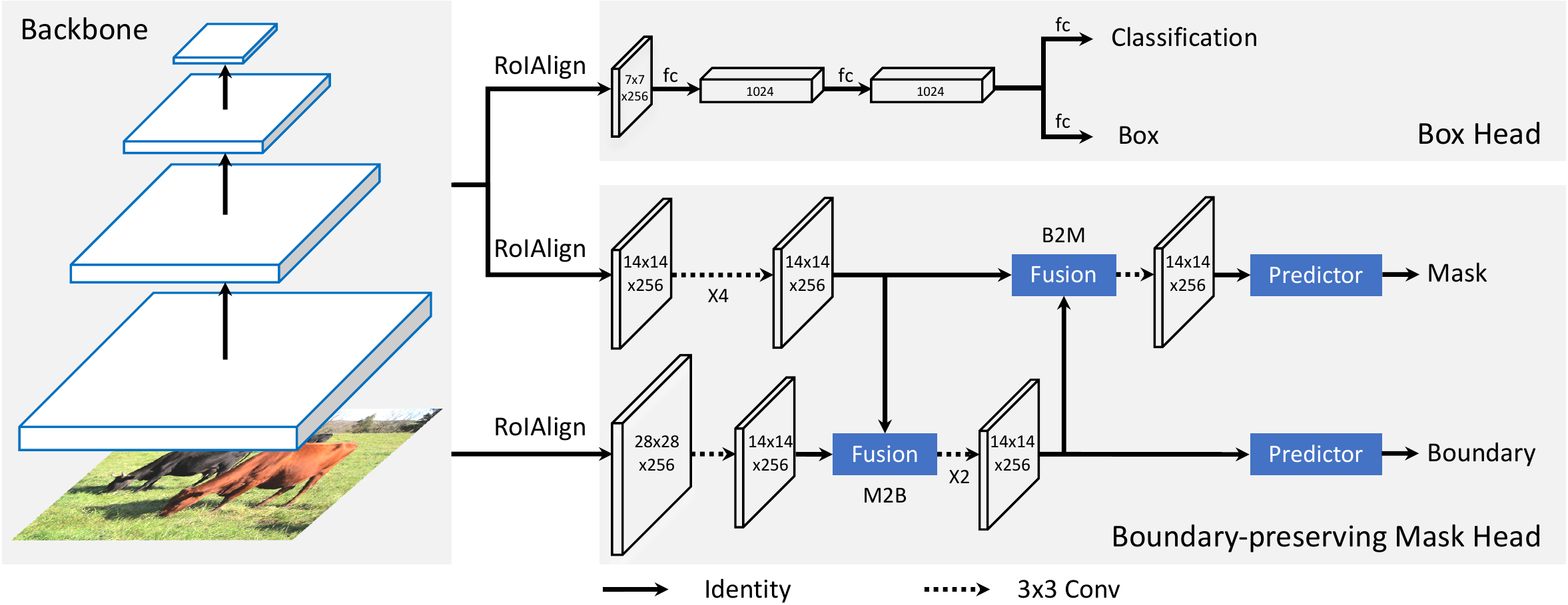}
    \caption{The Overall architecture of \textbf{Boundary-preserving Mask R-CNN (BMask R-CNN)}. The dotted arrow denotes $3\times3$ convolution and the solid arrow denotes identity connection unless specified annotation in boundary-preserving mask head.  ``$\times$4/$\times$2'' denotes a stack of four/two consecutive convs.}
    \label{fig:main_arch}
    \vspace{-0.4cm}
\end{figure}

\subsection{Boundary-preserving Mask Head}
\label{section-bgsh}

BMask R-CNN improves the mask head in Mask R-CNN with boundary features and boundary prediction, as illustrated in Fig.~\ref{fig:main_arch}. The new mask head is termed as boundary-preserving mask head. In the first stage, FPN extracts and constructs pyramidal features for RPN~\cite{RenHG017} and the heads in the second stage. After RPN generates proposals, the box head inherited from Mask R-CNN utilizes these proposals and extracted RoI features for classification and bounding box regression. The boundary-preserving mask head performs RoIAlign~\cite{HeGDG17} to acquire RoI features for both boundary and mask prediction. 

Boundary-preserving mask head jointly learns object boundaries and masks in an end-to-end manner. Note that object boundary and object mask have a close relation and we can easily convert either one to another. Features from the mask sub-network can provide high-level semantic information for learning boundaries. After obtaining boundaries, the shape information and abundant location information in boundary features can guide more precise mask predictions.  

\subsubsection{RoI Feature Extraction:}
We define $\mathcal{R}_m$ and $\mathcal{R}_b$ as Region of Interest (RoI) features for mask prediction and boundary prediction respectively. Following~\cite{LinDGHHB17}, $\mathcal{R}_m$ is extracted from the specific feature pyramid level ($P2 \sim P5$) according to the scale of the proposal, while $\mathcal{R}_b$ is obtained from the finest-resolution feature level $P2$, containing abundant spatial information. To preserve spatial information better for boundary prediction, the resolution of $\mathcal{R}_b$ is set to be larger than that of $\mathcal{R}_m$ when performing RoIAlgin. Then, it is downsampled by a strided $3\times3$ convolution and the output feature is denoted as $\widetilde{\mathcal{R}_b}$. $\widetilde{\mathcal{R}_b}$ has the same resolution as $\mathcal{R}_m$ and is used for feature fusion.

The feature fusion scheme in BMask R-CNN is illustrated in Fig.~\ref{fig:main_arch}. Mask RoI features $\mathcal{R}_m$ is fed into 4 consecutive $3\times3$ convolutions and the output feature is denoted as $\mathcal{F}_m$. Boundary features $\widetilde{\mathcal{R}_b}$ is fused with $\mathcal{F}_m$ and then fed into two consecutive $3\times3$ convolutions.
\vspace{-0.3cm}

\subsubsection{Mask $\rightarrow$ Boundary (M2B) Fusion:}
Mask features $\mathcal{F}_m$ contain rich high-level information, \ie, the pixel-wise object category information, which is beneficial to predict object boundaries. Hence, we propose a simple fusion block to integrate boundary features and mask features for boundary prediction. The fusion block can be formulated as follows.
\begin{equation}
    \mathcal{F}_b = f(\mathcal{F}_m) + \widetilde{\mathcal{R}_b}\text{,}
\end{equation}
where $\mathcal{F}_b$ denotes the output boundary features and $f$ means a $1\times1$ convolution with ReLU. 
\vspace{-0.3cm}

\subsubsection{Boundary $\rightarrow$ Mask (B2M) Fusion:}
We fuse the final boundary features with mask features; thus, boundary information can be used to enrich mask features and guide precise mask prediction. The fusion block is the same as the M2B fusion block.

\vspace{-0.3cm}
\subsubsection{Predictor:}
Following Mask R-CNN, our predictor is a $2\times2$ deconvolution followed by the $1\times 1$ convolution as output layer. Both masks and boundaries are class-specific.

\subsection{Learning and Optimization}

Following the common practice in edge detection~\cite{XieT17,YuF0R17}, we regard boundary prediction as a pixel-level classification problem. The learned boundary features are fused with mask features to provide shape information for mask prediction. 
\vspace{-0.3cm}
\subsubsection{Boundary Groundtruths:}

We use the Laplacian operator to generate soft boundaries from the binary mask groundtruths. The Laplacian operator is a second-order gradient operator and can produce thin boundaries. The produced boundaries are converted into binary maps by a threshold $0$ as the final groundtruths. 
\vspace{-0.7cm}

\subsubsection{Boundary Loss:}
Most boundary or edge detection methods~\cite{XieT17,YuF0R17,AcunaKF19} take the advantage of weighted cross-entropy to alleviate the class-imbalance problem in edge/boundary prediction. However, weighted binary cross-entropy leads to thick and coarse boundaries~\cite{DengSLWL18}. Following~\cite{DengSLWL18}, we use dice loss~\cite{MilletariNA16} and binary cross-entropy to optimize the boundary learning. Dice loss measures the overlap between predictions and groundtruths and is insensitive to the number of foreground/background pixels, thus alleviating the class-imbalance problem.
% However, Deng~\cite{DengSLWL18}~\etal indicate that weighted binary cross-entropy leads to thick and coarse boundaries and then exploit the combination of dice loss~\cite{MilletariNA16} and binary cross-entropy (BCE) loss. 
%Dice loss measures the overlap between predictions and groundtruths and is insensitive to the number of foreground/background pixels, thus alleviating the class-imbalance problem in boundary prediction.
% Following this practice, we use dice loss and binary cross-entropy to optimize the boundary learning, 
Our boundary loss $\mathcal{L}_b$ is formulated as follows.
\begin{equation}
\label{eq:lossb}
\mathcal{L}_{b}(p_b, y_b) =  \mathcal{L}_{Dice}(p_b, y_b) + \lambda \mathcal{L}_{BCE}(p_b, y_b)\text{,}
\end{equation}
where $H$ and $W$ are height and width of the predicted boundary map respectively, $p_b \in \mathbb{R}^{H\times W}$ denotes the predicted boundary for a particular category and $y_b \in \mathbb{R}^{H\times W}$ denotes the corresponding boundary groundtruth. $\lambda$ is a hyper-parameter to adjust the weight of dice loss (We set $\lambda = 1$ in all experiments). Dice loss is given as follows.
\begin{equation}
\mathcal{L}_{Dice}(p_b, y_b) = 1 - \frac{2\sum^{H\times W}_{i} p^i_by^i_b + \epsilon}{\sum^{H\times W}_i (p^i_b)^2 + \sum^{H\times W}_i (y^i_b)^2 + \epsilon}\text{,}
\end{equation}
where $i$ denotes the $i$-th pixel and $\epsilon$ is a smooth term to avoid zero division (We set $\epsilon=1$.). In ablation experiments, we will analyze and evaluate different loss functions with quantitative results and qualitative results.

\vspace{-0.3cm}
\subsubsection{Multi-Task Learning:}
Multi-task learning has been proved effective in many works~\cite{HeGDG17,DaiHS16,takikawa2019gated,KendallGC18,MisraSGH16}, which achieves better performance for different tasks comparing with separate training. Since boundary and mask are crossed linked by two fusion blocks, jointly training can enhance the feature representation for both boundary and mask predictions. We define a multi-task loss for each sample as follows. \begin{equation}
\mathcal{L} = \mathcal{L}_{cls} + \mathcal{L}_{box} + \mathcal{L}_{mask} + \mathcal{L}_{b}\text{,}
\end{equation}
where the classification loss $\mathcal{L}_{cls}$ and regression loss $\mathcal{L}_{box}$ are inherited from Mask R-CNN. Mask loss $\mathcal{L}_{mask}$ is a category-specific pixel-level binary cross-entropy
loss for each instance, which is directly taken from Mask R-CNN. The boundary loss $\mathcal{L}_{b}$ has been introduced in detail in Equation~\eqref{eq:lossb}. 

\section{Experiments}
We perform extensive experiments on the challenging COCO dataset~\cite{LinMBHPRDZ14} and the Cityscapes dataset~\cite{CordtsORREBFRS16} to demonstrate the effectiveness of Boundary-preserving Mask R-CNN. To better understanding each component of our method, we provide detailed ablation experiments on COCO. 

\vspace{-0.4cm}
\subsubsection{Dataset and Metrics:}
COCO contains 115k images for training, 5k images for validation and 20k images for testing. Our models are trained on the training set (\textit{train2017}). We report the results on the validation set (\textit{val2017}) for ablation studies and the results on testing set (\textit{test-dev2017}) to compare with other methods. The Cityscapes dataset is collected in urban scenes which contains 2975 training, 500 validation and 1525 testing images. As for instance segmentation, Cityscapes involves 8 object categories and provides more precise instance-level segmentation annotations than COCO. We train our models on the training set and report our performance on the validation set and the testing set. For both COCO and Cityscapes, we use the same evaluation metric (\ie, COCO AP), which is the average precision over different IoU thresholds (from 0.5 to 0.95). 
\vspace{-0.4cm}
\subsubsection{Implementation details:}
We adopt Mask R-CNN~\cite{massa2018mrcnn} as our baseline and our method is developed based on it. All hyper-parameters are kept the same. Unless specified, we use ResNet-50 with FPN as our backbone network. We initialize our backbone networks with ImageNet pre-trained weights and all Batch Normalization~\cite{IoffeS15} layers are frozen due to small mini-batch. The input images are resized such that the shorter side is 600 pixels and the longer is less than 1000 pixels for ablation experiments. As for main experiments, we adopt 800 pixels for shorter side (the longer is less than 1333 pixels). Following the standard practice, we train all models on 4 NVIDIA GPUs using Synchronized SGD with initial learning rate 0.02 and 16 images per mini-batch for 90,000 iterations and reduce the learning rate by a factor of 0.1 and 0.01 after 60,000 and 80,000 iterations respectively. For larger backbones, we follow the linear scaling rule~\cite{GoyalDGNWKTJH17} to adjust the learning rate schedule when decreasing mini-batch size.   

\subsection{Overall Results}
We first evaluate our BMask R-CNN with different backbones on COCO and compare it with Mask R-CNN. As shown in Table~\ref{tab:baseline_experiments}, our method outperforms Mask R-CNN by remarkable APs in spite of different backbones and input sizes. Compared with Mask R-CNN, BMask R-CNN significantly achieves $1.4$, $1.7$ and $1.5$ AP improvements using ResNet-50-FPN, ResNet-101-FPN and HRNetV2-W32-FPN~\cite{abs-1908-07919} respectively. Exploiting boundary information contributes to more precise mask localization due to the observation that our method yields noteworthy and stable improvements ($\approx 2.3$ AP) on AP$_{75}$. AP$^b$ shows AP for bounding box, on which BMask R-CNN very slightly improves over Mask R-CNN.

In Table~\ref{tab:coco_test_dev_results}, we compare BMask R-CNN with some state-of-the-art instance segmentation methods. All models are trained on COCO \textit{train2017} and evaluated on COCO \textit{test-dev2017}. Without bells and whistles, BMask R-CNN with ResNet-101-FPN can surpass these methods. 

Fig.~\ref{fig:ap_versus_iou} illustrates the AP curves of BMask R-CNN and Mask R-CNN under different IoU thresholds. Note that our method obtains larger gain when the IoU threshold increases, which shows the better localization performance of BMask R-CNN.
\vspace{-0.8cm}

\begin{table}[ht]
    \centering
    \caption{Comparison with Mask R-CNN on COCO \textit{val2017}}
    \vspace{0.2cm}
    \begin{tabular}{l|l|c|ccc|ccc}
    \toprule
    Method & Backbone & Input size & AP & AP$_{50}$ &  AP$_{75}$ & AP$^b$ & AP$^b_{50}$ &  AP$^b_{75}$ \\
    \midrule
    Mask R-CNN & ResNet-50-FPN    & 600 & 33.2 & 54.4 & 34.9 & 36.6 & 58.1 & 39.8 \\
    BMask R-CNN & ResNet-50-FPN   & 600 & \textbf{34.7} & 55.1 & \textbf{37.2} & 36.8 & 58.0 & 40.2 \\
    \hline
    Mask R-CNN & ResNet-50-FPN    & 800 & 34.2 & 56.0 & 36.3 & 37.8 & 59.2 & 41.1 \\
    BMask R-CNN & ResNet-50-FPN   & 800 & \textbf{35.6} & 56.3 & \textbf{38.4} & 37.8 & 59.0 & 41.5 \\
    \hline
    Mask R-CNN & ResNet-101-FPN   & 800 & 36.1 & 58.1 & 38.3 & 40.1 & 61.7 & 44.0 \\
    BMask R-CNN & ResNet-101-FPN  & 800 & \textbf{37.8} & 59.1 & \textbf{40.6} & 40.4 & 62.0 & 44.3 \\
    \hline 
    Mask R-CNN & HRNetV2-W32-FPN & 800 & 36.6 & 58.7 & 38.9 & 40.8 & 61.9 & 44.9 \\
    BMask R-CNN & HRNetV2-W32-FPN & 800 &  \textbf{38.1} & 59.4  & \textbf{40.7} &  41.0 & 61.9 & 45.1 \\
    \bottomrule
    \end{tabular}
    \label{tab:baseline_experiments}
\end{table}

\begin{table}[ht]
    \centering
    \caption{Comparison with state-of-the-art methods for instance segmentation on COCO \textit{test-dev2017}(* denotes our implementation)}
    \vspace{0.2cm}
    \begin{tabular}{l|l|ccc|ccc}
    \toprule
    Method & Backbone & AP & AP$_{50}$ &  AP$_{75}$  & AP$_{S}$ &  AP$_{M}$  &  AP$_{L}$ \\
    \midrule
    MNC~~\cite{DaiHS16}           & ResNet-101      & 24.6 & 44.3 & 24.8 & 4.7  & 25.9 & 43.6 \\
    FCIS~~\cite{LiQDJW17}         & ResNet-101      & 29.2 & 49.5 &  -   & -    & -    & -    \\
    FCIS+++~~\cite{LiQDJW17}      & ResNet-101      & 33.6 & 54.5 &  -   & -    & -    & -    \\
    Mask R-CNN~~\cite{HeGDG17}    & ResNet-101-FPN  & 35.7 & 58.0 & 37.8 & 15.5 & 38.1 & 52.4 \\
    Mask R-CNN~~\cite{HeGDG17}    & ResNeXt-101-FPN & 37.1 & 60.0 & 39.4 & 16.9 & 39.9 & 53.5 \\  
    MaskLab~~\cite{ChenHPS0A18}  & ResNet-101-FPN  & 35.4 & 57.4 & 37.4 & 16.9 & 38.3 & 49.2 \\
    MaskLab+~~\cite{ChenHPS0A18} & ResNet-101-FPN  & 37.3 & 59.8 & 39.6 & 19.1 & 40.5 & 50.6 \\
    Mask Scoring R-CNN~~\cite{HuangHGHW19} & ResNet-50-FPN & 35.8 & 56.5 & 38.4 & 16.2 & 37.4 & 51.0  \\
    Mask Scoring R-CNN~~\cite{HuangHGHW19} & ResNet-101-FPN & 37.5 & 58.7 & 40.2 & 17.2 & 39.5 & 53.0  \\
    CondInst~~\cite{Tian2020ConditionalCF} & ResNet-50-FPN & 35.4 & 56.4 & 37.6 & 18.4 & 37.9 & 46.9\\
    %CondInst$^{\dag}$~~\cite{Tian2020ConditionalCF} & ResNet-50-FPN & 37.8 & 59.1 & 40.5 & 21.0 & 40.3 & 48.7 \\
    BlendMask~~\cite{Chen2020BlendMaskTM} & ResNet-50-FPN & 34.3 & 55.4 & 36.6 & 14.9 & 36.4 & 48.9 \\
    %BlendMask$^{\dag}$~~\cite{Chen2020BlendMaskTM} & ResNet-50-FPN & 37.8 & 58.8 & 40.3 & 18.8 & 40.9 & 53.6 \\ 
    PointRend~~\cite{kirillov2019pointrend} & ResNet-50-FPN & 36.3 &  -   &  -   & -    &  -   &  -  \\
    \hline
    Mask R-CNN* & ResNet-50-FPN & 34.6 & 56.5 & 36.6 & 15.4 & 36.3 & 49.7 \\
    BMask R-CNN & ResNet-50-FPN & 35.9 & 57.0 & 38.6 & 15.8 & 37.6 & 52.2 \\
    \hline
    Mask R-CNN* & ResNet-101-FPN & 36.2 & 58.6 & 38.4 & 16.4 & 38.4 & 52.1 \\
    BMask R-CNN & ResNet-101-FPN & 37.7 & 59.3 & 40.6 & 16.8 & 39.9 & 54.6 \\
    BMask R-CNN w/ Mask Scoring & ResNet-101-FPN & 38.7 & 59.1 & 41.9 & 17.4 & 40.7 & 55.5 \\ 
    \bottomrule
    \end{tabular}
    \vspace{0.1cm}
    \label{tab:coco_test_dev_results}
    \vspace{-0.6cm}
\end{table}

\subsection{Ablation Experiments}
In order to comprehend how BMask R-CNN works, we perform exhaustive experiments to analyze the components in BMask R-CNN.
Table~\ref{tab:boundary_component_ablation} shows the results of gradually adding components to the Mask R-CNN baseline. 
Each component of our proposed BMask R-CNN will be investigated in the following sections.
\begin{table}[htb]
    \centering
    \small
    \caption{Experiment results on COCO \textit{val2017} of adding components to Mask R-CNN. 
    We gradually add boundary supervision (BCE loss), fusions between mask and boundary features, BCE-Dice loss 
    and our RoI feature extraction stategy for boundary features.}
    \vspace{0.2cm}
    \setlength{\tabcolsep}{2mm}{
    \begin{tabular}{cccc|ccc|c}
    \toprule
    Boundary & Fusions & BCE-Dice & RoI Strategy  & AP & AP$_{50}$ & AP$_{75}$ &  AP$^b$ \\
    \midrule
        -      &      -     &      -     &     -      & 33.2 & 54.4 & 34.9 & 36.6 \\
    \checkmark &            &            &            & 33.9 & 54.8 & 35.8 & 36.7 \\
    \checkmark & \checkmark &            &            & 34.2 & 55.4 & 36.4 & 36.8 \\
    \checkmark & \checkmark & \checkmark &            & 34.4 & 55.0 & 36.6 & 36.7 \\
    \checkmark & \checkmark & \checkmark & \checkmark & 34.7 & 55.1 & 37.2 & 36.8 \\
    \bottomrule
    \end{tabular}}
    \vspace{-0.6cm}
    \label{tab:boundary_component_ablation}
\end{table}
 
\vspace{-0.4cm}
\subsubsection{Effects of Boundaries:}
To validate the effect of boundaries for mask prediction, 
we use mask targets to replace boundary targets and also evaluate the performance without bounadry supervision and loss with the architecture kept the same.
Table~\ref{tab:effect_of_boundary} indicates that boundary supervision with our proposed boundary-preserving mask head improves mask results by $0.8$ and $0.7$ AP compared with mask supervision and no supervision respectively.
Notably, using boundary can improve the mask localization performance (AP$_{75}$) by a significant margin.
\begin{table}[htb]
    \centering
    \caption{Experiment results on COCO \textit{val2017} of changing groundtruth of the boundary head. 
    \xmark~denotes no supervision on the boundary head.}
    \vspace{0.2cm}
    % \vspace{0.1cm}
    \small
    \setlength{\tabcolsep}{3mm}{
    \begin{tabular}{l|ccc|c}
    \toprule
    Groundtruth  & AP & AP$_{50}$ & AP$_{75}$ & AP$^b$ \\
    \midrule
    \xmark   & 34.0 & 55.0 & 36.0 & 36.3 \\
    mask     & 33.9 & 54.3 & 35.9 & 36.0 \\
    boundary & 34.7 & 55.1 & 37.2 & 36.8 \\
    \bottomrule
    \end{tabular}}
    \label{tab:effect_of_boundary}
    \vspace{-0.2cm}
\end{table}

\vspace{-0.4cm}
\subsubsection{RoI Feature Extraction:}
\begin{figure}
    \begin{minipage}[c]{0.4\linewidth} 
        \centering
        \captionof{table}{Experiment results on COCO \textit{val2017} for different RoI feature extraction strategies.}
        \vspace{-0.2cm}
        \footnotesize
        \begin{tabular}{l|c|ccc|c}
            \toprule
            Source & Size & AP & AP$_{50}$ & AP$_{75}$ & AP$^b$ \\
            \midrule
            P2 $\sim$ P5 & 14 & 34.4 & 55.1 & 36.5 & 36.8 \\
            P2           & 14 & 34.5 & 55.0 & 36.7 & 36.7 \\    
            \hline
            P2 $\sim$ P5 & 28 & 34.4 & 55.1 & 36.8 & 36.6 \\
            P2           & 28 & 34.7 & 55.1 & 37.2 & 36.8 \\
            \bottomrule
        \end{tabular}
        % \vspace{-0.5cm}
        \label{tab:boundary_roi_feature}
    \end{minipage}
    \begin{minipage}[c]{0.58\linewidth}
        
        \subfigure[]{
            \centering
            \includegraphics[width=0.46\linewidth]{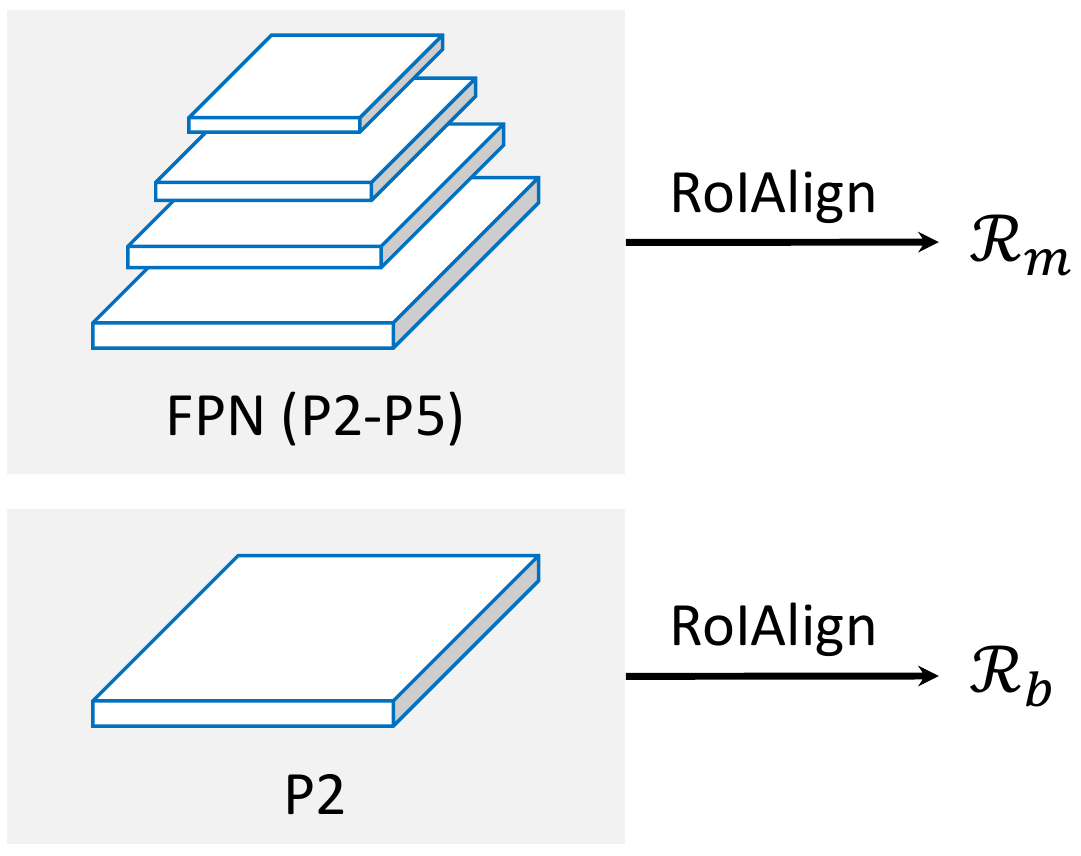}
            \label{fig:feature_extraction_a}
        }
        \subfigure[]{
            \centering
            \includegraphics[width=0.46\linewidth]{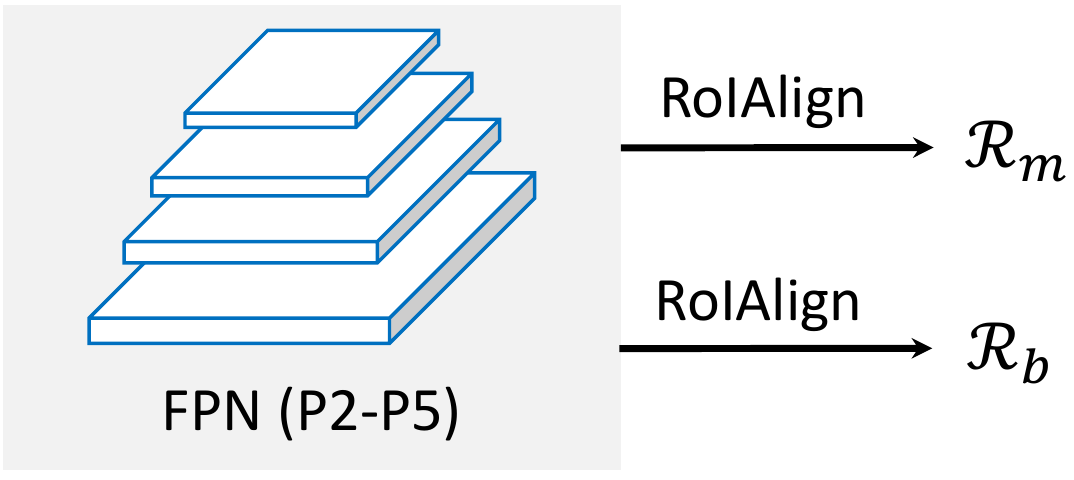}
            \label{fig:feature_extraction_b}
        }
    \end{minipage}
    \begin{minipage}[t]{\linewidth}
        \captionof{figure}{Different RoI feature extraction strategies for Boundary-preserving Mask Head.}
        \label{fig:featrure_extraction}
    \end{minipage}
    \vspace{-0.9cm}
\end{figure}

Compared with mask prediction, predicting boundaries requires more precise spatial information due to boundaries are spatially sparse. Therefore, we explore several strategies and present two considerations to extract better RoI features for boundaries. The first aspect is the source of RoI features. Lin~\etal~\cite{LinDGHHB17} propose that RoI features are extracted from the different levels ($P2\sim P5$) in FPN depending on the scales of the corresponding proposals. Features of high levels in FPN lacks spatial information which are inappropriate for boundaries.
Fig.~\ref{fig:featrure_extraction} illustrates different sources for mask features $\mathcal{R}_m$ and boundary features $\mathcal{R}_b$. Fig.~\ref{fig:feature_extraction_a} shows that boundary features are directly extracted from $P2$ while mask features are from from $P2\sim P5$ according to the scale of the proposal. Fig.~\ref{fig:feature_extraction_b} shows that both boundary features and mask features are extracted from the same feature level from $P2\sim P5$.
%Fig.~\ref{fig:featrure_extraction} illustrates the sources for mask features $\mathcal{R}_m$ and boundary features $\mathcal{R}_b$, in which boundary features are directly extracted from $P2$ while mask features are from from $P2\sim P5$ according to the scale of the proposals.
The other aspect is the feature resolution. Higher-resolution features preserve more spatial information which is beneficial to boundaries. Therefore, we explore the effects of $28\times28$ resolution and $14\times14$ resolution RoI features for learning boundaries. 
As Table~\ref{tab:boundary_roi_feature} shows, directly extracting boundary RoI features from $P2$ is more effective with larger resolution. We employ boundary features extracted from $P2$ with $28\times28$ resolution in other experiments.

\vspace{-0.4cm}
\subsubsection{Feature Fusion:}
In Section~\ref{section-bgsh}, we have emphasized the relation between boundary features and mask features. Fusion blocks in our boundary-preserving mask head build explicit links to enrich both feature representation. 
Table~\ref{tab:fusion} shows more results: if there is no fusion, it improves Mask R-CNN by $0.5$ AP, which is the gain of multi-task learning; with both M2B and B2M fusion blocks, BMask R-CNN has $1.5$ AP improvement over Mask R-CNN. We further investigate the influence of adding more subsequent fusion blocks. Keeping the overall computation cost substantially unchanged, adding more B2M or M2B fusions brings negligible improvements. 
\vspace{-0.3cm}

\begin{table}[ht]
    \centering
    \caption{Experiment results on COCO \textit{val2017} for the impacts of fusion blocks, \ie, mask $\rightarrow$ boundary (M2B) fusion and boundary $\rightarrow$ mask (B2M) fusion.}
    \vspace{0.2cm}
    \small
    \setlength{\tabcolsep}{1.8mm}{
    \begin{tabular}{cc|ccc|c}
    \toprule
    M2B Fusion & B2M Fusion & AP & AP$_{50}$ & AP$_{75}$ & AP$^{b}$ \\
    \midrule
    \xmark      & \xmark     & 33.7 & 55.0 & 35.7 & 36.8 \\
    \checkmark  & \xmark     & 34.2  & 54.9 & 36.5 & 36.8 \\    
    \xmark      & \checkmark & 33.9 & 54.7 & 36.1 & 36.6 \\
    \checkmark  & \checkmark & 34.7 & 55.1 & 37.2 & 36.8 \\
    \bottomrule
    \end{tabular}}
    \label{tab:fusion}
    \vspace{-0.2cm}
\end{table}

\subsubsection{Loss Functions:}
In order to obtain precise boundaries, we evaluate the impacts of different loss functions for optimizing boundary learning. Table~\ref{tab:loss_and_optimization} shows that the combination of BCE and Dice loss leads to better performance compared with individual BCE or Dice loss. Weighted BCE brings less gain than BCE in boundary prediction.

To investigate how the Dice-BCE combined loss provides such competitive improvements, we present detailed analysis on the visualization results of these experiments. 
As shown in Fig.~\ref{fig:loss_comparison}, different loss functions have different impacts on learning boundaries. BCE loss provides considerably precisely-localized but unclear boundaries due to the class-imbalance problem. Weighted BCE solves this problem by applying balancing weights but this hard balancing leads to thick and coarse boundaries which exceed their corresponding masks. Dice loss also solves the class-imbalance problem without thick boundaries but lacks precise localization. Consequently, combining Dice loss and BCE can provide better-localized boundaries and avoid the class-imbalance problem.
\vspace{-0.3cm}

\begin{table}
\begin{minipage}[b]{0.45\linewidth} 
    \centering
    \captionof{table}{Experiment results on COCO \textit{val2017} for evaluating different loss functions.}
    \footnotesize
    \begin{tabular}{l|ccc|c}
    \toprule
    Loss type  & AP & AP$_{50}$ & AP$_{75}$ & AP$^{b}$\\
    \midrule
    BCE          & 34.3 & 54.9 & 36.3 & 36.7 \\
    Weighted BCE & 34.1 & 55.0 & 36.2 & 36.7 \\    
    Dice         & 34.3 & 55.0 & 36.5 & 36.5 \\
    \hline
    Dice-BCE     & 34.7 & 55.1 & 37.2 & 36.8 \\
    \bottomrule
    \end{tabular}
    \vspace{0.5mm}
    \label{tab:loss_and_optimization}
\end{minipage}
\begin{minipage}[b]{0.53\linewidth}
    \centering
    \includegraphics[width=0.9\linewidth]{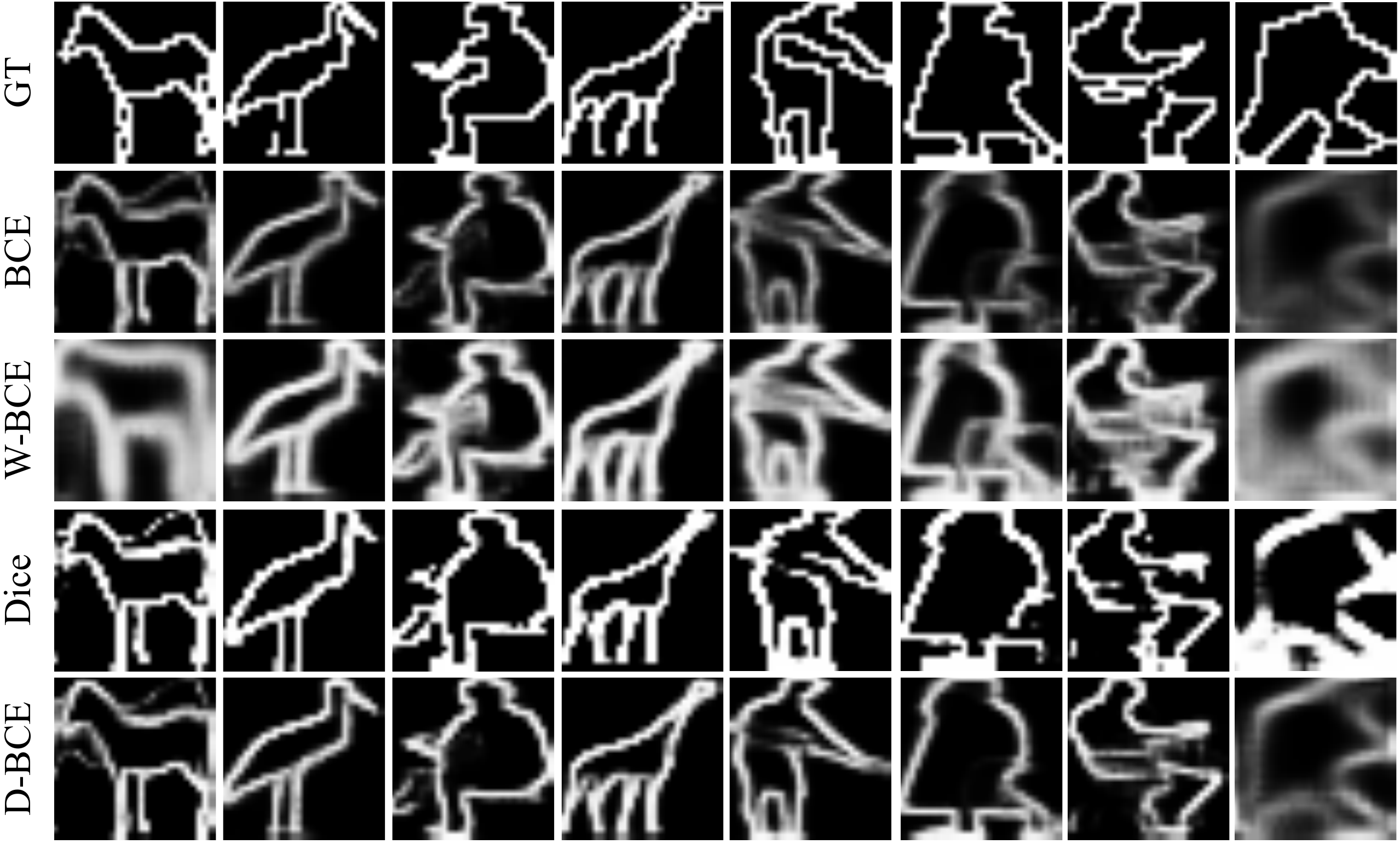}
\end{minipage}
\begin{minipage}{\linewidth}
    \captionof{figure}{Visualization results for analyzing the impacts of different loss functions. \textbf{GT} denotes groundtruth. \textbf{W-BCE} and \textbf{D-BCE} denote Weighted BCE and Dice-BCE respectively.}
    \label{fig:loss_comparison}
\end{minipage}
\vspace{-0.6cm}
\end{table}

\subsubsection{Computation Cost:}
Compared with Mask R-CNN, our method involves four $3\times3$ and two $1\times1$ convolutional layers for boundary prediction and two fusion blocks which increase the computation cost. To clarify the improvements of BMask R-CNN are not from extra computation cost, we form a larger mask head by adding 4 more $3\times3$ convolutional layers as a comparison. Table~\ref{tab:comparison_headvier_head} shows that BMask R-CNN still achieves a significant gain compared with Mask R-CNN with equal computation cost.
\vspace{-0.3cm}
\begin{table}[ht]
    \centering
    \caption{Experiment results on COCO \textit{val2017} for evaluating the impacts of computation cost. 
    MRCNN and LMH denote Mask R-CNN and the larger mask head respectively. 
    FLOPs are counted only for mask head without the final predictors. 
    Inference time is tested on one NVIDIA RTX 2080Ti with the input size 600*1000}
    \vspace{0.2cm}
    \footnotesize
    \setlength{\tabcolsep}{1.8mm}{
    \begin{tabular}{l|cc|ccc|c}
    \toprule
    Method & FLOPs & Time(ms/img.) & AP & AP$_{50}$ & AP$_{75}$ & AP$^{b}$\\
    \midrule
    MRCNN         & 0.46G & 59.0 & 33.2 & 54.4 & 34.9 & 36.6 \\
    MRCNN w/~ LMH & 0.93G & 61.3 & 33.7 & 54.6 & 35.8 & 36.7 \\
    \hline
    BMask R-CNN   & 0.95G & 63.7 & 34.7 & 55.1 & 37.2 & 36.8 \\
    \bottomrule
    \end{tabular}}
    \vspace{-0.5cm}
    \label{tab:comparison_headvier_head}
\end{table}

\begin{table}[ht]
    \centering
    \caption{Experiment results on Cityscapes \texttt{val} (AP[\texttt{val}]) and \texttt{test}.(* denotes our implementation)}
    \vspace{0.2cm}
    \small
    \scriptsize
    \begin{tabular}{l|cc|cc|cccccccc}
    \toprule
    & AP~[\texttt{val}] & AP$_{75}$~[\texttt{val}] & AP & AP$_{50}$ & person & rider & car & truck & bus & train & mcycle & bicycle \\
    \midrule
    % InstanceCut~~\cite{KirillovLASR17} &   -  & 13.0 & 27.9 & - & - & - & - & - & - & - & -  \\
    BAIS~~\cite{HayderHS17}     &  -   &   -  & 17.4 & 36.7 & - & - & - & - & - & - & - & -  \\ 
    DIN~~\cite{ArnabT17} &  -   &   -  & 20.0 & 38.8 & 16.5 & 16.7 & 25.7 & 20.6 & 30.0 & 23.4 & 17.1 & 10.1 \\
    % SGN~~\cite{LiuJFU17} & \texttt{fine} + \texttt{coarse} & 29.2 & 25.0 & 44.9 & 21.8 & 20.1 & 39.4 & 24.8 & 33.2 & 30.8 & 17.7 & 12.4 \\
    SGN~~\cite{LiuJFU17}       & 29.2 &   -   & 25.0 & 44.9 & 21.8 & 20.1 & 39.4 & 24.8 & 33.2 & 30.8 & 17.7 & 12.4 \\
    Mask R-CNN~~\cite{HeGDG17} & 31.5 &  -   &  26.2 & 49.9 & 30.5 & 23.7 & 46.9 & 22.8 & 32.2 & 18.6 & 19.1 & 16.0 \\
    BshapeNet~~\cite{abs-1810-10327}  &  -  &  -   &  27.1 & 50.3 & 29.6 & 23.3 & 46.8 & 25.8 & 32.9 & 24.6 & 20.3 &  14.0 \\
    BshapeNet+~~\cite{abs-1810-10327} &  -  &  -   &  27.3 & 50.5 & 30.7 & 23.4 & 47.2 & 26.1 & 33.3 & 24.8 & 21.5 & 14.1 \\
    Neven~\etal~~\cite{NevenBPG19} &  -  &  -   &  27.6 & 50.9 & 34.5 & 26.1 & 52.4 & 21.7 & 31.2 & 16.4 & 20.1 & 18.9 \\
    \hline
    Mask R-CNN* & 32.0  &  30.1  &  27.2 & 53.0 & 31.4 & 23.7 & 49.1 & 22.9 & 33.7 & 21.9 & 19.4 & 15.4 \\
    BMask R-CNN        & 35.0 &  33.6   & 29.4 & 54.7 & 34.3 & 25.6 & 52.6 & 24.2 & 35.1 & 24.5 & 21.4 & 17.1 \\
    \bottomrule
    \end{tabular}
    \label{tab:cityscapes_results}
    \vspace{-0.5cm}
\end{table}

\subsection{Experiments on Cityscapes}
To further explore the effects of BMask R-CNN on the fine-annotated Cityscapes dataset, we only use images with \texttt{fine} annotations to train and evaluate our models. For fair comparisons, we use ResNet-50-FPN as our backbone and resize images with shorter edge randomly selected from [800, 1024] for training. For inference, input images are kept the original size 1024$\times$2048. Models are trained by SGD on 4 GPUs with mini-batch size 4 for 48,000 iterations. The learning rate is 0.005 at the beginning and reduced to 0.0005 after 36,000 iterations. Other settings are the same with experiments on COCO. 

We report the results evaluated on Cityscapes \texttt{val} and \texttt{test} in Table~\ref{tab:cityscapes_results}. BMask R-CNN achieves $29.4$ AP on \texttt{test} and obtains a remarkable $2.2$ AP gain compared with the baseline Mask R-CNN. BMask R-CNN outperforms previous methods without extra data. 

\subsection{Discussions}

\subsubsection{Coarse boundary annotation \vs precise boundary prediction:} 
When datasets become larger and larger, obtaining precise mask annotations is unavoidably time-consuming. 
Though the COCO dataset provides abundant instance-level annotations, the mask and boundary annotations (represented by sparse polygons) are coarse, which limits the performance of our method BMask R-CNN. 
Nevertheless, BMask R-CNN can output more precise and smooth boundaries with fewer mask overlap between instances; some selected examples are shown in Fig.~\ref{fig:coco_annotation_prediction}.

\vspace{-0.4cm}
\subsubsection{Sobel mask head:} 
\begin{figure}[htb]
    \centering
    \subfigure[]{
        \centering
        \includegraphics[width=0.5\linewidth]{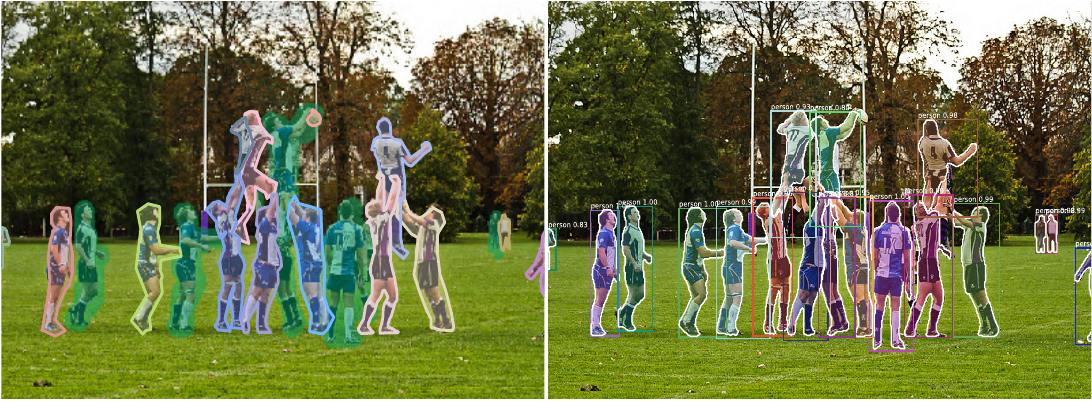}
        \label{fig:coco_annotation_prediction}
    }
    \subfigure[]{
        \centering
        \includegraphics[width=0.40\linewidth]{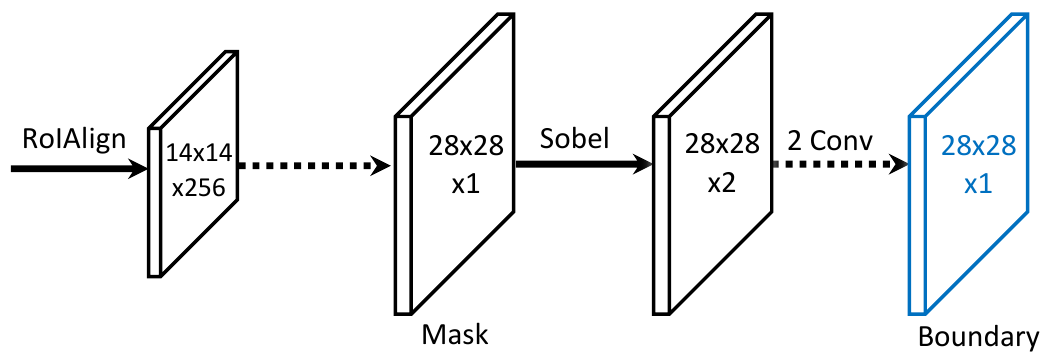}
        \label{fig:sobel_mask_head}
    }
    \caption{
        \textbf{(a):} Qualitative comparison between COCO \textit{val} annotations (left) and our instance boundary predictions (right).
        \textbf{(b):} \textbf{Sobel Mask Head:} We use Sobel operator to obtain 2-channel boundary features indicate $X$ direction and $Y$ direction and then apply two $3\times3$ convolutions to output boundary predictions.}
    \vspace{-0.5cm}
\end{figure}

Instead of predicting boundaries using an extra branch, we also design a simple Sobel mask head to predict boundaries from masks, 
which is a improved version of~\cite{abs-1809-07069}.
As illustrated in Fig.~\ref{fig:sobel_mask_head}, it has a Sobel operator~\cite{Kittler83} and two $3\times3$ convolutions following the mask predictions. 
We adopt the same Dice-BCE loss function for training.
% Sobel operator is less sensitive to noise compared to Laplacian operator.
Using ResNet-50-FPN backbone and keep the rest settings the same, this Sobel mask head method obtains $34.0$ AP which improves Mask R-CNN by $0.8$ AP 
but is $0.7$ AP worse than our main method.
\vspace{-0.5cm}
% It obtains 34.0 AP on COCO with ResNet-50-FPN backbone and input size of 600$\times$1000.

\subsection{Qualitative Results}
\begin{figure}
    \subfigure[]{
        \centering
        \includegraphics[width=0.60\linewidth]{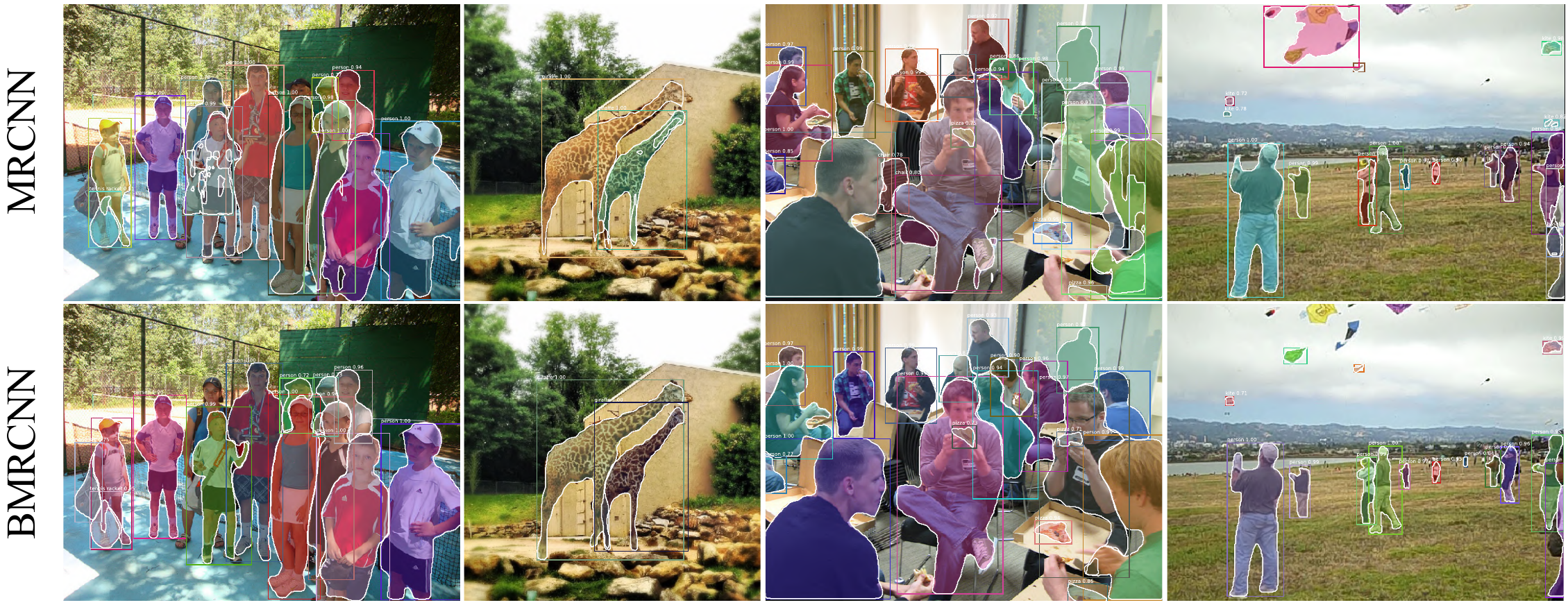}
        \label{fig:qualitative_coco}
    }
    \subfigure[]{
        \centering
        \includegraphics[width=0.37\linewidth]{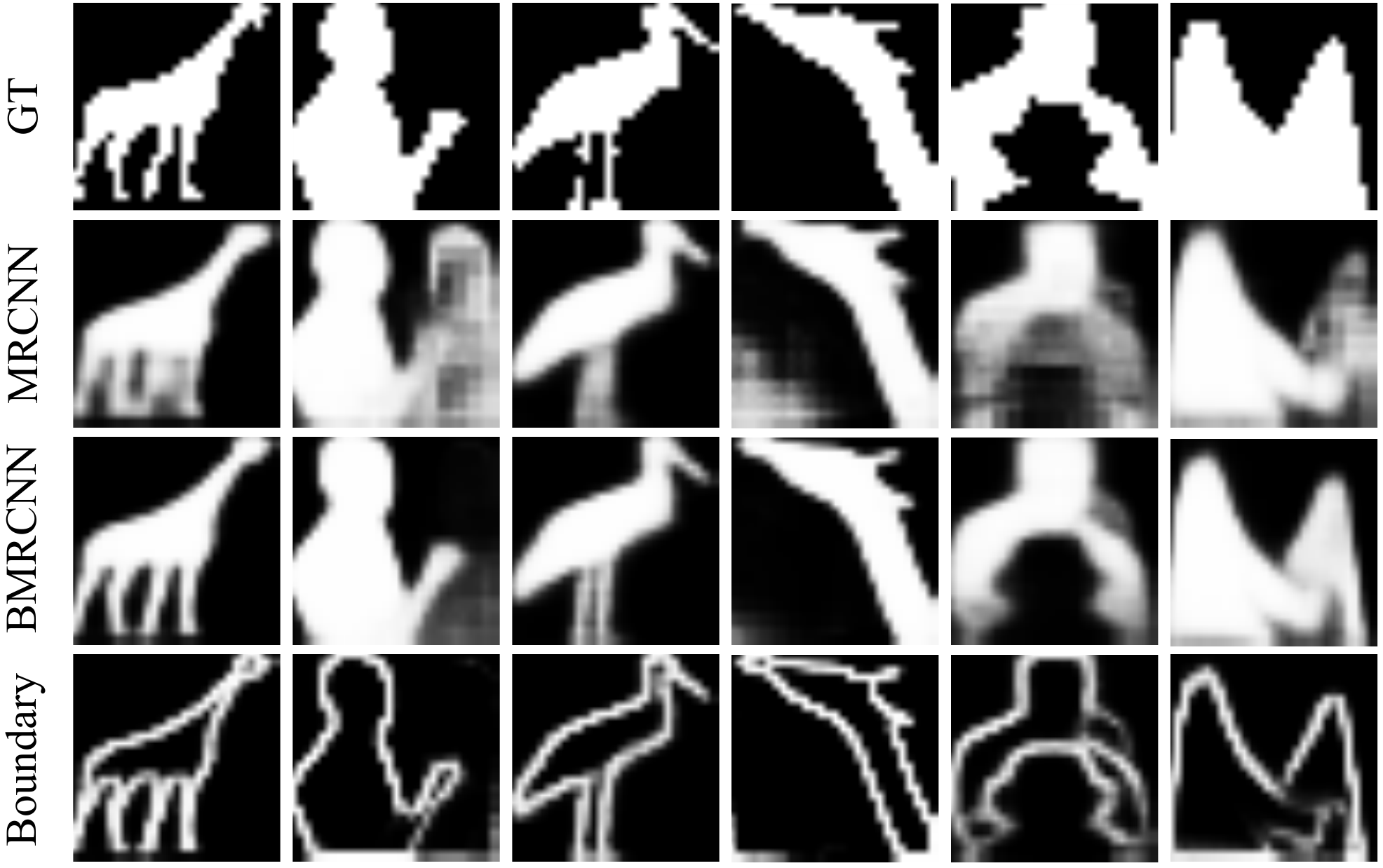}
        \label{fig:qualitative_raw_coco}
    }
    \caption{\textbf{(a):} Qualitative results on COCO dataset generated by Mask R-CNN and BMask R-CNN with ResNet-101-FPN. MRCNN and BMRCNN denotes Mask R-CNN and BMask R-CNN respectively. \textbf{(b):} Raw mask prediction and boundary prediction from boundary-preserving mask head. GT: the groundtruth segmentation. MRCNN: mask predicted by Mask R-CNN. BMRCNN: mask predicted by BMask R-CNN. Boundary: boundary predicted by BMask R-CNN. Results are obtained with ResNet-101-FPN backbone.}
    \vspace{-0.5cm}
\end{figure}

We provide representative visualization results on COCO to compare our method with Mask R-CNN and further prove the effectiveness of our method.
% Visualize COCO
Fig.~\ref{fig:qualitative_coco} shows the qualitative results on COCO \textit{val}. Mask R-CNN is more prone to generate masks with coarse boundaries which contain much background along with some false positive areas. Our proposed BMask R-CNN can alleviate this issue with the help of preserving boundaries.
% Visualize boundary and segmentation 
We further visualize our raw boundary and mask results in Fig.~\ref{fig:qualitative_raw_coco}. It can be easily observed that predicted masks are more clear and highly coincident with their boundaries. Furthermore, utilizing predicted boundaries to refine masks brings minor improvement and the refinement is vulnerable to the noises.

\begin{comment}
\begin{figure}[htb]
    \centering
    \includegraphics[width=\linewidth]{figures/cocovis_compressed.pdf}
    \caption{Qualitative results on COCO dataset generated by Mask R-CNN and BMask R-CNN with ResNet-101-FPN. MRCNN and BMRCNN denotes Mask R-CNN and BMask R-CNN respectively.}
    \label{fig:qualitative_coco}
    \vspace{-0.8cm}
\end{figure}

\begin{figure}[htb]
    \centering
    \includegraphics[width=0.6\linewidth]{figures/coco-raw.pdf}
    \caption{Raw mask prediction and boundary prediction from boundary-preserving mask head. \textbf{GT}: the groundtruth segmentation. \textbf{MRCNN}: mask predicted by Mask R-CNN. \textbf{BMRCNN}: mask predicted by BMask R-CNN. \textbf{Boundary}: boundary predicted by BMask R-CNN. Results are obtained with ResNet-101-FPN backbone.}
    \label{fig:qualitative_raw_coco}
    \vspace{-0.4cm}
\end{figure}

\end{comment}

\section{Conclusion}
We address the issue that coarse boundaries and imprecise localization in instance segmentation and propose a novel Boundary-preserving Mask R-CNN. 
It incorporates boundary information to guide the mask learning for better boundaries and localization. Our experiments demonstrate that our method achieves remarkable and stable improvements on both COCO and Cityscapes especially in terms of localization performance. 
Extensive studies and visualization results provide a deep understanding of how our method BMask R-CNN works.
Our method could also be plugged into Cascade Mask R-CNN and \etc for higher performance. We hope it can be a strong baseline and sheds light on this fundamental research topic.

\section*{Acknowledgements}
This work was in part supported by NSFC (No. 61733007 and No. 61876212), Zhejiang Lab (No. 2019NB0AB02), and HUST-Horizon Computer Vision Research Center.

\bibliographystyle{splncs04}
\bibliography{egbib}
\end{document}